\documentclass[11pt]{article}

\usepackage[final]{acl}

\usepackage{times}
\usepackage{latexsym}
\usepackage[T1]{fontenc}
\usepackage[utf8]{inputenc}
\usepackage{microtype}
\usepackage{graphicx}
\usepackage{booktabs}
\usepackage{multirow}
\usepackage{amsmath}
\usepackage{xcolor}
\usepackage{float}
\usepackage{pifont}
\newcommand{\yes}{\ding{51}}
\newcommand{\no}{\ding{55}}

\newcommand{\nepooc}{\textsc{NepOOC}}

\title{\nepooc-M: Bilingual Nepali-English Benchmark and Comparative Analysis of Multimodal Architectures for OOC Detection}

\author{Sanjeev Khatiwada \\
  Independent Researcher, Kathmandu, Nepal \\
  \texttt{skhatiwada558@gmail.com}}

\begin{document}
\maketitle

\begin{abstract}
Out-of-context (OOC) misinformation pairs authentic images with misleading captions to construct false narratives without image manipulation, making detection a problem of multimodal alignment rather than image forensics. Despite the prevalence and consequences of OOC misinformation in Nepal, no public benchmark exists for Nepali. We introduce \nepooc, the first publicly available Nepali-dominant multilingual OOC benchmark, comprising 1,090 image--caption pairs (545 pristine, 545 OOC) annotated across five typologies (fabricated, miscaptioned, temporal mismatch, geographic mismatch, identity mismatch) with inter-annotator agreement $\kappa = 0.84$. Systematic evaluation of five multimodal architectures alongside text-only and image-only baselines shows that text-only models perform strongly on the current benchmark: a text-only mBERT model reaches 94.65 $\pm$ 0.20\% Macro-F1, with no statistically detectable difference from the best multimodal system (ResNet-50+mBERT, 94.56 $\pm$ 0.24\%). Image-only models perform near chance, as expected because the two captions of each item share one image. Caption-only baselines show that terminal punctuation and caption length separate the classes to a large degree, and retraining the neural models on captions without punctuation lowers Macro-F1 by about 21 points. We therefore interpret the reported scores as caption discrimination on this paired benchmark, and recommend text-only and caption-style baselines as a standard check for OOC benchmarks.
\end{abstract}

\section{Introduction}

Out-of-context (OOC) misinformation exploits authentic, unmanipulated images by pairing them with false or misleading captions to construct plausible-looking but semantically inaccurate claims \citep{luo2021newsclippings,aneja2021cosmos,papadopoulos2024verite}. Unlike synthetic or doctored imagery, OOC content passes visual plausibility checks and is consequently difficult for both humans and automated systems to detect \citep{abdelnabi2022open}. The core challenge is not whether the image is authentic, but whether the image and caption together accurately represent a real event or situation.

In Nepal, OOC misinformation is both prevalent and consequential, spanning elections, public health, natural disasters, and communal affairs. A representative case involved a Madhesh Province official who used photographs of Delhi to falsely claim provincial infrastructure development \citep{dahal2025anatomy}. In another incident, an old photograph was paired with an inflammatory caption tied to the 2023 Dharan controversy, nearly inciting communal conflict before fact-checkers intervened \citep{adhikari2025fake}. These cases share a common structure that distinguishes Nepali OOC from English-language counterparts: the critical evidence lies in caption claims about regional actors, locations, and events that require localised knowledge to evaluate.

Despite the severity of this problem, OOC detection research remains concentrated in high-resource, English-language settings. Benchmarks such as NewsCLIPpings \citep{luo2021newsclippings}, COSMOS \citep{aneja2021cosmos}, and VERITE \citep{papadopoulos2024verite} have driven methodological progress, yet none addresses challenges unique to a low-resource, bilingual environment: code-switching between Devanagari and Latin scripts, sparse named-entity coverage for Nepali figures and locations, and the absence of multimodal OOC tools adapted to regional contexts. We refer to this as the \emph{regional context gap}: a vision-language model that performs well on English news may still fail on Nepali content because it lacks the script-level and cultural grounding needed to identify image--caption inconsistencies in this domain.

This paper addresses the regional context gap through three contributions:
\begin{enumerate}
\item[1)] \textbf{\nepooc{} Dataset.} The first publicly available Nepali-dominant multilingual OOC benchmark, comprising 1,090 image--caption pairs annotated across five typologies with Cohen's $\kappa = 0.84$, drawn from real-world Nepali misinformation incidents.
\item[2)] \textbf{Comparative Architecture Study.} Systematic evaluation of five multimodal architectures with dedicated text-only and image-only baselines, designed to identify which inductive biases and modalities transfer effectively to this low-resource, bilingual setting.
\item[3)] \textbf{Text-Only Strength and Caption-Cue Analysis.} Multi-metric evaluation indicates that text-only models perform strongly within this benchmark. Because the two captions of each item share one image, image-only chance follows from the benchmark construction, and caption-only baselines together with a punctuation ablation show that terminal punctuation and caption length separate the classes to a large degree.
\end{enumerate}

\section{Related Work}

\subsection{Out-of-Context Misinformation Detection}

OOC misinformation differs from image forgery or deepfake detection: the image is authentic, but its caption displaces it from its original context \citep{luo2021newsclippings,aneja2021cosmos}.

Early work established the task infrastructure. NewsCLIPpings \citep{luo2021newsclippings} introduced automatic pipelines for constructing large-scale OOC benchmarks from news corpora. COSMOS \citep{aneja2021cosmos} proposed self-supervised consistency objectives for grounded image--caption representations, while VERITE \citep{papadopoulos2024verite} incorporated unimodal-bias controls to ensure models perform genuine cross-modal reasoning. Alongside these benchmarks, entity-enhanced fusion \citep{qi2021improving}, transformer-based architectures \citep{yang2023multimodal}, and context-aware reasoning \citep{zhang2023interpretable} improved detection in high-resource settings, and open-domain methods extended the task by retrieving web evidence \citep{abdelnabi2022open}. Modality contribution studies \citep{eldien2026interpreting} document that text often provides sufficient signal in certain settings, a finding this work corroborates for \nepooc{} while showing that, here, it is largely a property of the benchmark construction.

\subsection{Vision-Language Architectures and Fusion Strategies}

On the visual side, convolutional networks, particularly ResNet \citep{he2016deep}, remain competitive baselines for spatial feature extraction, while Vision Transformers (ViT) \citep{dosovitskiy2021image} extend patch-based representation learning to visual inputs. CLIP \citep{radford2021learning} yields highly transferable image--text representations via large scale contrastive pretraining, though this objective optimises global similarity ranking rather than supervised binary classification, which may limit applicability to small, domain-specific benchmarks.

On the text side, BERT-style encoders \citep{devlin2019bert} and their multilingual extensions \citep{conneau2020unsupervised} provide strong cross-lingual transfer baselines. Fusion strategies span late concatenation, cross-attention over patch tokens, and graph-based relational modelling \citep{kananian2023gramufen}. A recurring observation across the literature, and one we also see in our ablations, is that architectural complexity does not consistently improve performance when labelled data are scarce; strong pretrained text encoders often suffice \citep{eldien2026interpreting}.

\subsection{Low-Resource and Nepali Language Processing}

Work on Nepali NLP has grown, though multimodal misinformation detection remains underexplored. MuRIL \citep{khanuja2021muril}, trained on 17 Indian and South Asian languages including Devanagari, provides the strongest off-the-shelf foundation for Nepali text processing. Nepali-specific language models, NepBERTa \citep{timilsina2022nepberta} and NepaliBERT \citep{pudasaini2023nepalibert}, together with transformer-based classifiers \citep{maskey2022nepali,shahi2018nepali,wagle2021comparative,adhikari2023nepali} demonstrate that script-aware pretraining improves text classification. These models, however, address text-only tasks and do not extend to multimodal or OOC settings.

At the multimodal frontier, recent Indic-language work has begun combining MuRIL with Vision Transformers for multilingual fake news detection \citep{vedaksha2026multimodal}, though without a dedicated OOC evaluation or typology schema.

No prior work provides a multimodal OOC benchmark with typology-level annotations in Nepali or any closely related low-resource South Asian language that we are aware of. This is the gap that \nepooc{} fills.

\begin{table}[t]
\centering\small
\setlength{\tabcolsep}{3pt}
\begin{tabular}{@{}lccccr@{}}
\toprule
Benchmark & OOC & Typol. & Multi. & L-R & Size \\
\midrule
NewsCLIPpings & \yes & \no & \no & \no & 71k \\
COSMOS & \yes & \no & \no & \no & 200k \\
VERITE & \yes & \no & P & \no & 1k \\
MuMiN & \no & \no & \yes & P & 22k \\
\nepooc{} (ours) & \yes & \yes & \yes & \yes & 1,090 \\
\bottomrule
\end{tabular}
\caption{Benchmark comparison. \yes=yes, \no=no, P=partial. Typol.=typology labels; Multi.=multilingual; L-R=low-resource. MuMiN: \citet{nielsen2022mumin}.}
\label{tab:bench}
\end{table}

\subsection{Positioning \nepooc{}}

Table~\ref{tab:bench} situates \nepooc{} relative to the most relevant existing benchmarks. \nepooc{} is the first resource to simultaneously offer OOC-specific focus, fine-grained typology labels ($\kappa = 0.84$), genuine bilingual content reflecting real code-switching patterns, and pairs drawn from actual Nepali misinformation incidents.

\section{\nepooc{} Dataset}

\nepooc{} reflects three core design principles: all pairs are drawn from real-world misinformation instances (ecological validity); each OOC instance carries a typology annotation (typological granularity); and the benchmark encompasses Nepali, English, and code-switched captions (multilingual realism).

\subsection{Data Collection and Sources}
\label{sec:collection}

\nepooc{} comprises 545 unique image--source pairs, each yielding one pristine and one OOC image--caption pair, for a total of 1,090 samples. The two samples of a pair share one \texttt{post\_id} and one image URL, so each image appears once with a pristine caption and once with an OOC caption. The dataset covers a broad range of domains in Nepali digital media, including public affairs, elections, health crises, natural disasters, infrastructure claims, and social controversy. Sources were drawn from three categories: fact-checking organisations operating in Nepal, Nepali online news portals, and social media archives flagged for potential misinformation. For all source types, we documented both the false claim and the verified true context from the same evidentiary reference, enabling direct derivation of OOC and pristine labels without post-hoc re-annotation. Fact-checking organisations constitute the dominant source (469 of 545 items, 86.1\%; the rest are 49 social-media archive items and 27 news-portal items), reflecting a deliberate prioritisation of expert-verified ground truth over scale. We acknowledge this introduces a source bias toward professionally curated content; see Section~\ref{sec:limits}. For the fact-checker items, the Pristine and OOC captions were copied verbatim from the fact-checking article, and debunk wording that appears in the article was retained: explicit debunk phrases (for example ``misleading'', ``claim is false'' and their Nepali equivalents; phrase list in the repository) occur in 89 of 545 OOC captions (16.3\%) and in 164 of 545 Pristine captions (30.1\%). For the remaining 76 items, the annotators wrote the captions from the verified claim and context.

\subsection{Annotation Protocol and Inter-Annotator Agreement}

Binary OOC labels for fact-checker-sourced pairs were established by the fact-checkers themselves as part of their published verdicts, constituting expert-verified ground truth. For pairs sourced from news portals and social media archives ($n = 76$; 14.0\% of unique sources), two trained annotators confirmed the binary OOC/Pristine classification (Cohen's $\kappa = 0.81$), and all OOC instances received one of five typology labels (Section~\ref{sec:typology}) using published fact-checker analyses as reference material (Cohen's $\kappa = 0.84$). Because this subset is small, the annotators worked through the claims and their verified contexts together and then passed the completed entries to a further reviewer for approval; we report the resulting agreement values as measured by the annotators, and the individual annotation sheets were not retained.

\subsection{OOC Typology Schema}
\label{sec:typology}

Each OOC instance is annotated with one of five typology labels reflecting distinct classes of media manipulation observed in Nepali discourse:
\begin{enumerate}
\item[1)] \textbf{Fabricated}: the caption describes an event or situation that did not occur; the authentic image lends false visual credibility to the invented claim.
\item[2)] \textbf{Miscaptioned}: the caption is factually inaccurate but does not involve a systematic mismatch of time, location, or identity.
\item[3)] \textbf{Temporal mismatch}: an authentic image from one time period is recirculated with a caption implying a different, typically more recent, time.
\item[4)] \textbf{Geographic mismatch}: an authentic image captured at one location is presented as depicting a different location.
\item[5)] \textbf{Identity mismatch}: an authentic image of one person or group is captioned as depicting a different person or group.
\end{enumerate}
Fabricated instances are the largest category (54.9\%), reflecting the prevalence of entirely invented captions in Nepali misinformation. Identity mismatch is the rarest subtype (1.8\%, $n = 10$), creating substantial within-OOC class imbalance.

\begin{table}[t]
\centering\small
\begin{tabular}{@{}llr@{}}
\toprule
Attribute & Category & Count (\%) \\
\midrule
\multirow{2}{*}{Label} & Pristine & 545 (50.0\%) \\
 & OOC & 545 (50.0\%) \\
\midrule
\multirow{3}{*}{Language} & Nepali & 856 (78.5\%) \\
 & English & 158 (14.5\%) \\
 & Code-switched & 76 (7.0\%) \\
\midrule
\multirow{3}{*}{Split} & Train & 754 (69.2\%) \\
 & Validation & 108 (9.9\%) \\
 & Test & 228 (20.9\%) \\
\midrule
\multicolumn{2}{@{}l}{Typology IAA (Cohen's $\kappa$)} & 0.84 \\
\multicolumn{2}{@{}l}{Binary IAA, non-FC (Cohen's $\kappa$)} & 0.81 \\
\bottomrule
\end{tabular}
\caption{Dataset statistics. Percentages are of all 1,090 samples. IAA=inter-annotator agreement; non-FC=not from a fact-checking organisation.}
\label{tab:stats}
\end{table}

\subsection{Dataset Statistics and Splits}
\label{sec:splits}

The final benchmark contains 1,090 image--caption pairs partitioned into train/validation/test splits (754/108/228) using a pair-grouped split: both samples of a \texttt{post\_id} are always in the same split (377/54/114 items for train/validation/test), and no \texttt{post\_id} or image URL occurs in more than one split. Table~\ref{tab:stats} summarises the full dataset statistics. Captions are predominantly Nepali (78.5\%), with English (14.5\%) and code-switched (7.0\%) varieties.

Nine image URLs are shared by several \texttt{post\_id}s (545 items map to 527 URLs). In 27 items (4 in the test set) the two captions are identical once punctuation and whitespace are removed, so no caption-only model can label both samples of such an item correctly. All 545 images were matched to their \texttt{post\_id} and decoded; SHA-256 hashing finds 9 groups of identical files, none across splits. A perceptual-hash search (Hamming distance $\le$ 8) returns 27 candidate pairs, 13 of them across splits; one train/validation pair is the same photograph in two sizes, and the others show different photographs that share a fact-checker card template. A review of the 38 items labelled Nepali whose captions are mostly in Latin script suggests 36 English, 1 code-switched and 1 uncertain; the released labels are unchanged. All checks are listed in Appendix~\ref{app:integrity}.

\subsection{Leakage Checks}
\label{sec:leakage}

We did not perform an event-level leakage analysis, so we cannot rule out that items reporting the same real-world event fall on both sides of the train--test split. Leakage is controlled at the level of pairs and images: both captions of an item always share a split, and the URL, file-hash and perceptual-hash checks of Section~\ref{sec:splits} found no exact duplicates across splits and one resized duplicate between the training and validation sets. The consequences for interpreting our results are discussed in Section~\ref{sec:limits}.

\section{Methodology}

\subsection{Task Formulation and Preprocessing}

Given an image--caption pair $(I, T)$, the task is binary classification into Pristine (label 0) or OOC (label 1). The training objective minimises standard binary cross-entropy loss over $N$ training samples, where $y_i \in \{0,1\}$ is the ground truth label and $\hat{p}_i$ is the model's predicted probability for class 1. Images are resized to $224\times224$ ($448\times448$ for ViT+MuRIL) and channel normalised using ImageNet statistics. No data augmentation is applied. Captions are tokenised using each model's native tokeniser: mBERT (\texttt{bert-base-multilingual-cased}; vocabulary 119,547), MuRIL (\texttt{google/muril-base-cased}; vocabulary 197,285), and CLIP's BPE tokeniser. All tokenisations are padded or truncated to a maximum sequence length of 128 tokens.

\begin{figure*}[t]
\centering
\includegraphics[width=0.66\textwidth]{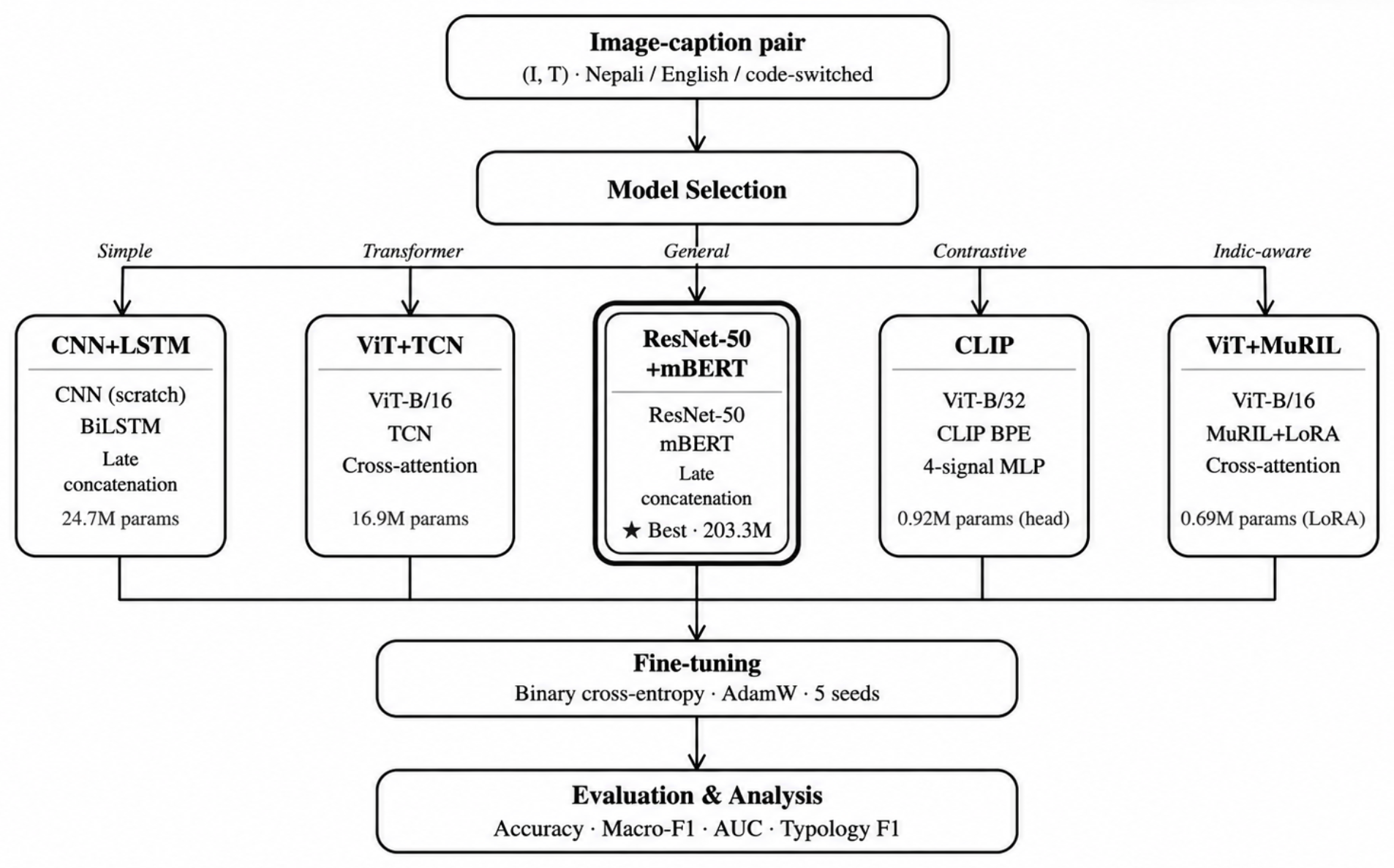}
\caption{Five multimodal architectures organised by visual encoder, text encoder, and fusion mechanism. Double border indicates the best-performing multimodal model (ResNet-50+mBERT; 94.56\% Macro-F1). Parameter counts are trainable parameters (see Table~\ref{tab:arch}).}
\label{fig:arch}
\end{figure*}

\subsection{Model Selection and Architecture Overview}

Five models are selected across three distinct architectural dimensions: (i) visual encoder type, from scratch CNN (CNN+LSTM), supervised convolutional pretraining (ResNet-50), contrastive vision-language pretraining (CLIP ViT-B/32), and supervised ViT pretraining (ViT-B/16); (ii) text encoder type, generic multilingual BERT (mBERT), Devanagari-aware multilingual BERT with LoRA adaptation (MuRIL; \citealp{hu2022lora}), and lightweight temporal convolution (TCN, included for stable convergence under limited data; \citealp{bai2018empirical}); and (iii) fusion mechanism, late concatenation (CNN+LSTM, ResNet-50+mBERT), cross-attention over patch tokens (ViT+TCN, ViT+MuRIL), and similarity-based head (CLIP). This design enables direct comparison of architectural choices under identical splits, training protocols, and evaluation metrics. Figure~\ref{fig:arch} provides a schematic overview of all five architectures.

\subsection{Architecture Specifications}

Table~\ref{tab:arch} summarises key architectural specifications. CNN+LSTM uses randomly initialised embeddings and bidirectional LSTM (256 dimensions per direction). ViT+TCN employs single-head cross-attention over ViT patch tokens. ResNet-50+mBERT uses separate learning rates for visual ($1\times10^{-4}$) and text ($2\times10^{-5}$) encoders. CLIP keeps both encoders frozen, training only a classification head. ViT+MuRIL applies LoRA with rank $r = 8$, $\alpha = 16$ to MuRIL, adding 0.29M LoRA parameters (0.69M trainable in total, including the cross-attention and classifier). The text-only baselines are trained separately: mBERT is fully fine-tuned (AdamW, learning rate $2\times10^{-5}$, weight decay $10^{-4}$, batch size 32, cosine schedule with warm-up), while text-only MuRIL uses the same LoRA setting as ViT+MuRIL (learning rate $2\times10^{-5}$, weight decay 0.05). All models are trained on the 754/108/228 splits across 5 independent random seeds $\{42, 123, 456, 789, 2024\}$ with early stopping applied based on validation Macro-F1 with patience 10--12 epochs depending on the model. The seeds change initialisation, data order and training subsets, not the split. The initial experiments required approximately 120 GPU-hours on NVIDIA T4 GPUs; the re-runs of the two focal models reported in Sections~\ref{sec:main} to \ref{sec:typ} required about 4 GPU-hours. Full training configurations are in Appendix~\ref{app:config}.

\begin{table*}[t]
\centering\footnotesize
\begin{tabular}{@{}llllll@{}}
\toprule
Model & VE & TE & Fusion & TE Vocab & Params \\
\midrule
CNN+LSTM & 5-layer CNN (512) & LSTM (256/dir) & Late concat & 119k & 24.7M \\
ViT+TCN & ViT-B/16 (frozen) & TCN (3 blocks) & Cross-attention & 119k & 16.9M \\
ResNet-50+mBERT & ResNet-50 (2048$\to$768) & mBERT (768) & Late concat & 119k & 203.3M \\
CLIP & ViT-B/32 (frozen) & CLIP BPE (frozen) & Similarity head & 49k & 0.92M (head) \\
ViT+MuRIL & ViT-B/16 (frozen) & MuRIL+LoRA (r=8) & Cross-attention & 197k & 0.69M (LoRA) \\
\bottomrule
\end{tabular}
\caption{Architecture specifications. VE=visual encoder, TE=text encoder, Params=trainable parameters; ViT+TCN has 103.5M with the frozen ViT, ViT+MuRIL LoRA adds 0.29M, ResNet-50+mBERT fine-tunes all layers.}
\label{tab:arch}
\end{table*}

\section{Experimental Setup}

\subsection{Evaluation Metrics}

Accuracy, Macro-F1, and AUC (area under the ROC curve) are reported across all experiments, with Macro-F1 as the primary metric. OOC-class precision and recall are additionally reported, derived from per-seed average confusion matrices. All metrics are computed on the held-out test set ($n = 228$; 114 Pristine, 114 OOC), averaged across 5 random seeds with standard deviation (ddof $= 1$). For the best-performing models, the exact (binomial) McNemar test is applied to per-seed predictions matched by sample, and the Macro-F1 difference is estimated with a paired bootstrap that resamples \texttt{post\_id} pairs (both captions together; 10,000 replicates) and averages over seeds.

\subsection{Training-Size Scaling Experiment}

Text-only mBERT and ResNet-50+mBERT are trained on 25\%, 50\%, 75\% and 100\% of the 377 training pairs. Pairs are sampled as units, so both captions of a \texttt{post\_id} stay together; validation and test sets are fixed, and each fraction is repeated over 5 seeds.

\subsection{Caption-Cue Analysis}
\label{sec:audit}

To quantify how much of the task can be solved from the caption alone, we train caption-only baselines on length features, style features (terminal punctuation, danda, quotation marks, digits, script ratios and similar), and word- and character-level TF-IDF with logistic regression or a linear SVM. Vectorisers and scalers are fit on the training set, regularisation strength is selected by validation Macro-F1, the test set is scored once, and confidence intervals resample \texttt{post\_id}. Each baseline is also evaluated after removing all Unicode punctuation and symbol characters and collapsing whitespace; the two focal neural models are retrained on the same normalised captions with the original recipe and five seeds. A publisher hold-out, which trains without one publisher domain and tests on it, serves as a diagnostic of source dependence, and a trained ResNet-50+mBERT is probed with shuffled and black test images to test whether its image branch is used.

\section{Results}

\subsection{Main Benchmark Results}
\label{sec:main}

Table~\ref{tab:main} reports benchmark results averaged across 5 seeds, including dedicated text-only baselines. Text-only mBERT and ResNet-50+mBERT reach 94.65 $\pm$ 0.20 and 94.56 $\pm$ 0.24 Macro-F1. The two systems were compared with the exact McNemar test on matched samples: $p = 1.000$ in all five seeds, with 0.6 discordant samples per seed on average. The paired bootstrap difference (ResNet-50+mBERT minus mBERT) is $-$0.09 pp, 95\% CI [$-$0.44, 0.18]. No statistically detectable difference was observed under these test conditions; with so few discordant samples the test has little power, so this is not evidence of equivalence. An earlier set of runs gave the reverse ordering (94.56 and 94.65, respectively); neither ordering is distinguishable. All results use the pair-grouped random split (Section~\ref{sec:splits}).

\begin{table*}[t]
\centering\small
\begin{tabular}{@{}llccc@{}}
\toprule
Model & Type & Acc. (\%) & Macro-F1 (\%) & AUC \\
\midrule
mBERT & Text-only & 94.65 $\pm$ 0.20 & 94.65 $\pm$ 0.20 & 0.9797 $\pm$ 0.0052 \\
MuRIL & Text-only & 94.38 $\pm$ 0.35 & 94.38 $\pm$ 0.35 & 0.9567 $\pm$ 0.0158 \\
\midrule
ResNet-50+mBERT & Multimodal & 94.56 $\pm$ 0.24 & 94.56 $\pm$ 0.24 & 0.9654 $\pm$ 0.0081 \\
ViT+MuRIL & Multimodal & 93.33 $\pm$ 0.37 & 93.33 $\pm$ 0.37 & 0.9505 $\pm$ 0.0057 \\
ViT+TCN & Multimodal & 92.11 $\pm$ 1.35 & 92.10 $\pm$ 1.36 & 0.9616 $\pm$ 0.0064 \\
CNN+LSTM & Multimodal & 78.16 $\pm$ 10.97 & 78.15 $\pm$ 10.97 & 0.8548 $\pm$ 0.1203 \\
CLIP & Multimodal & 69.39 $\pm$ 0.72 & 69.00 $\pm$ 0.72 & 0.7127 $\pm$ 0.0028 \\
\bottomrule
\end{tabular}
\caption{Main results on test split ($n = 228$, mean$\pm$std over 5 seeds). mBERT and ResNet-50+mBERT: re-runs with saved predictions; other rows: original runs.}
\label{tab:main}
\end{table*}

\subsection{OOC-Class Precision and Recall}

Table~\ref{tab:ooc} (Appendix~\ref{app:ooc}) extends the main results with OOC-class precision and recall. Text-only mBERT and ResNet-50+mBERT have the same OOC Precision (95.37 and 95.36\%); mBERT's OOC Recall is 93.86\% in all five seeds (standard deviation exactly 0.00). CLIP shows severely imbalanced precision--recall (Precision 74.97\%, Recall 58.25\%), indicating systematic misclassification of OOC instances as Pristine.

\subsection{Training-Size Scaling}

Table~\ref{tab:scale} reports Macro-F1 at each training fraction for the two re-run models. Both are close to their final score with 25\% of the training pairs (94 pairs): mBERT reaches 94.47 $\pm$ 0.67 and ResNet-50+mBERT 93.86 $\pm$ 0.82, against 94.65 $\pm$ 0.20 and 94.56 $\pm$ 0.24 with all 377 pairs. A style/punctuation logistic regression trained on the same fractions scores 92.5 at each fraction, and a character TF-IDF SVM 85.8, 89.4, 91.0 and 91.7 at 25, 50, 75 and 100\%. Early saturation is therefore consistent with a caption cue that is available from very few training pairs (Section~\ref{sec:shortcut}).

\begin{table*}[t]
\centering\small
\begin{tabular}{@{}lcccc@{}}
\toprule
Model & 25\% (94 pairs) & 50\% (188) & 75\% (283) & 100\% (377) \\
\midrule
mBERT (text-only) & 94.47 $\pm$ 0.67 & 94.38 $\pm$ 0.65 & 94.56 $\pm$ 0.50 & 94.65 $\pm$ 0.20 \\
ResNet-50+mBERT & 93.86 $\pm$ 0.82 & 94.56 $\pm$ 0.39 & 94.56 $\pm$ 0.39 & 94.56 $\pm$ 0.24 \\
\bottomrule
\end{tabular}
\caption{Scaling results: Macro-F1 (mean$\pm$std over 5 seeds). Pairs (both captions) are sampled as units; validation and test sets are fixed.}
\label{tab:scale}
\end{table*}

\subsection{Caption-Cue Results}
\label{sec:shortcut}

Pristine captions are longer than their OOC partner in 80.4\% of items (77.2\% in the test set; mean 107.6 versus 71.8 characters), and 97.8\% of Pristine captions end in a danda or full stop against 10.8\% of OOC captions. Table~\ref{tab:audit} shows the consequence. Length alone gives 68.3 Macro-F1 and word TF-IDF 66.8, but a logistic regression on punctuation and script features reaches 92.5 (95\% CI 89.0--95.6) and a character n-gram SVM 91.7, against 94.65 for mBERT; once punctuation is removed, these fall to 66.0 and 66.9. Retraining the neural models on captions without punctuation (same recipe, split and seeds) lowers Macro-F1 from 94.65 to 73.90 $\pm$ 1.74 for mBERT and from 94.56 to 72.58 $\pm$ 1.66 for ResNet-50+mBERT, drops of 20.75 pp (95\% CI [16.01, 25.81]) and 21.98 pp ([17.03, 27.30]) on matched seeds. The neural models stay above the punctuation-free baselines (66--67), so some signal remains, but the image does not restore the loss: the two models differ by $-$1.31 pp (95\% CI [$-$3.28, 0.59]) and no seed gives a significant exact McNemar test (median $p = 0.50$). Most of the benchmark score is therefore not recovered once punctuation is removed. Holding out a publisher does not remove the effect (Appendix~\ref{app:pub}): the character SVM reaches 93.5 on \texttt{nepalfactcheck.org} and 78.8 on \texttt{techpana.com}, and falls to about 62 on both once punctuation is removed.

\begin{table}[t]
\centering\scriptsize
\setlength{\tabcolsep}{2pt}
\begin{tabular}{@{}lccc@{}}
\toprule
Model (caption only) & Macro-F1 & 95\% CI & No punct. \\
\midrule
Length-only LR & 68.3 & 63.0--73.6 & 67.3 \\
Style/punctuation LR & 92.5 & 89.0--95.6 & 66.0 \\
Word TF-IDF LR & 66.8 & 61.0--72.6 & 66.8 \\
Char TF-IDF SVM & 91.7 & 88.1--95.2 & 66.9 \\
\midrule
mBERT (text-only) & 94.65 $\pm$ 0.20 & --- & 73.90 $\pm$ 1.74 \\
ResNet-50+mBERT & 94.56 $\pm$ 0.24 & --- & 72.58 $\pm$ 1.66 \\
\bottomrule
\end{tabular}
\caption{Caption-only baselines on the released split (228 test rows, 114 pairs). CI: bootstrap over pairs. Neural rows: mean $\pm$ std over 5 seeds. All baselines: Appendix~\ref{app:base}.}
\label{tab:audit}
\end{table}

\section{Analysis}

\subsection{Modality Ablation}

Table~\ref{tab:abl} reports Macro-F1 for text-only, image-only, and multimodal configurations. Image-only performance falls at or below chance (33.33--49.83\%) across all models; this is expected, because the same image appears with both labels (Section~\ref{sec:collection}). Text-only configurations approach or match multimodal performance: ViT+MuRIL's text-only result (93.95 $\pm$ 0.84\%) surpasses its multimodal counterpart (93.33 $\pm$ 0.37\%) by 0.62 pp, and ResNet-50+mBERT does not gain from visual features (94.56 versus 94.65 for text-only mBERT).

To check whether the image branch is used at all, trained ResNet-50+mBERT models were evaluated with test images swapped between pairs and with black images. Macro-F1 stays between 93.9 and 94.7 in every seed, and at most 1 of 228 predictions (0.44\%) changes in any seed, although the image branch receives gradient during training. The trained model therefore does not depend on the image on this benchmark.

\begin{table*}[t]
\centering\small
\begin{tabular}{@{}lccc@{}}
\toprule
Model & Text-only (\%) & Image-only (\%) & Multimodal (\%) \\
\midrule
CNN+LSTM & 74.91 $\pm$ 16.80 & 49.80 $\pm$ 0.27 & 78.15 $\pm$ 10.97 \\
ViT+TCN & 90.87 $\pm$ 1.80 & 33.33 $\pm$ 0.00 & 92.10 $\pm$ 1.36 \\
CLIP & 67.48 $\pm$ 1.06 & 49.20 $\pm$ 0.48 & 69.00 $\pm$ 0.72 \\
ResNet+mBERT & 94.65 $\pm$ 0.20 & 49.00 $\pm$ 0.96 & 94.56 $\pm$ 0.24 \\
ViT+MuRIL & 93.95 $\pm$ 0.84 & 49.83 $\pm$ 0.18 & 93.33 $\pm$ 0.37 \\
\bottomrule
\end{tabular}
\caption{Modality ablation: Macro-F1 at 100\% training data (mean$\pm$std over 5 seeds).}
\label{tab:abl}
\end{table*}

\subsection{Confusion Matrix Analysis and Typology-Level Performance}
\label{sec:typ}

Models perform strongly on dominant typologies: ResNet-50+mBERT recognises 95.7\% of Miscaptioned and 91.9\% of Fabricated OOC test captions (mean over seeds). Due to severe class imbalance, Temporal, Geographic, and Identity mismatches are grouped as ``Other Mismatches'' ($n = 24$); individual Identity mismatch results are not reliable. As a separate five-way task on OOC captions only, character TF-IDF reaches 57.6 Macro-F1 and word TF-IDF 43.0, and a fine-tuned mBERT 50.95 $\pm$ 3.80 (five seeds; per-class F1: Fabricated 0.80, Miscaptioned 0.54, Temporal 0.62, Geographic 0.59, Identity 0.00; Appendix~\ref{app:typ}). Identity mismatch has only 7, 1 and 2 captions in train, validation and test, and the five-way results are exploratory.

\subsection{Precision--Recall and ROC Curves}

The divergence between AUC and Macro-F1 for CLIP (AUC = 0.7127, F1 = 69.00\%) illustrates the importance of reporting both metrics; AUC alone would overstate CLIP's suitability. Text-only mBERT has the higher AUC (0.9797 $\pm$ 0.0052 against 0.9654 $\pm$ 0.0081 for ResNet-50+mBERT).

\subsection{Failure Case Analysis}

ResNet-50+mBERT makes 12--13 errors per seed (5.3--5.7\% error rate) and 14 distinct test samples are misclassified in at least one seed: 8 missed OOC captions (Fabricated 5, Miscaptioned 2, Geographic mismatch 1) and 6 Pristine captions flagged as OOC. Twelve samples are misclassified in every seed, as by text-only mBERT, and two of them belong to pairs whose captions are identical once punctuation is removed. All failures share a common root cause: the closed-pair constraint prevents access to external fact-checking, knowledge bases, and visual grounding beyond the dataset scale.

\section{Discussion}

\subsection{Why Multimodal Gains Remain Limited}

Image-only baselines perform at or below chance (33--50\%) across all architectures. Because both captions of an item share one image, the image alone carries no label information, so this is guaranteed by construction and does not show that visual features are uninformative in general. Adding visual information to text-only models yields no measurable benefit here: text-only mBERT and ResNet-50+mBERT differ by 0.09 percentage points (94.65 versus 94.56), with a bootstrap interval that includes zero, and text-only ViT+MuRIL surpasses its multimodal variant by 0.62 pp. This holds for both concatenation-based fusion (ResNet) and cross-attention (ViT+MuRIL), but testing fusion would need a benchmark in which image content decides the label. The caption cue in Section~\ref{sec:shortcut} is a second reason why text alone does well.

\subsection{Dataset Scale vs.\ Architecture}

At 754 training samples, pretrained text encoders already reach about 94 Macro-F1 with a quarter of the training pairs. Whether larger, more diverse training sets or architectural changes would help more cannot be decided from this benchmark, since saturation is also consistent with the caption cue.

A secondary observation is that LoRA-adapted MuRIL shows no advantage over fully fine-tuned mBERT. Both text-only encoders differ by only 0.27 pp, and the multimodal ViT+MuRIL trails ResNet-50+mBERT by 1.23 pp. The two encoders are tuned differently (LoRA versus full fine-tuning), so this is not a clean comparison of script-specific and generic pretraining. The closed-pair constraint (no external knowledge or evidence retrieval) limits performance on minority typologies: Temporal, Geographic, and Identity mismatches comprise 20\% of OOC cases but require world knowledge unavailable within the image--caption pair alone. Fabricated claims (55\% of OOC) similarly demand external fact-checking. Addressing these challenges requires retrieval-augmented architectures, as demonstrated by recent work (RED-DOT, \citealp{papadopoulos2024reddot}; EXCLAIM, \citealp{wu2025exclaim}).

\section{Limitations}
\label{sec:limits}

\paragraph{Dataset scale.} At 1,090 samples, \nepooc{} is substantially smaller than high-resource benchmarks. A single misclassification on the 228-sample test set shifts Macro-F1 by approximately 0.4 percentage points, limiting statistical power to detect small differences. Typology-level findings for Identity mismatch ($n = 2$ OOC test samples) are not reliable.

\paragraph{Shared-image pairing and text-dominant signal.} Each image appears with both a pristine and an OOC caption, so image-only chance and text-only/multimodal parity are expected by construction, limiting conclusions about cross-modal fusion.

\paragraph{Source and caption-style bias.} Sources are concentrated among fact-checking organisations (86.1\%) covering salient events, introducing skew toward professionally curated content. Pristine and OOC captions differ systematically in length and punctuation (Section~\ref{sec:shortcut}), and removing punctuation lowers the neural Macro-F1 by about 21 pp, so high overall performance partly reflects style patterns specific to this source distribution. The publisher hold-out is a diagnostic only, not external validation, and we know of no independent Nepali OOC dataset on which to test.

\paragraph{Event-level leakage.} Splits are grouped by pair and screened for duplicate images, but items describing the same event may still appear in both train and test, which could inflate scores.

\paragraph{Image content and provenance.} Many images are screenshots of fact-checker cards that carry a verdict stamp and the publisher's logo rather than the original photograph, and one test image (BBC\_20170901\_1) is a $480\times2$ pixel placeholder. Fact-checker captions retain the debunk wording of the source article, and the agreement statistics rest on the annotators' own records. The logs of the initial experiments were not retained: we re-ran mBERT and ResNet-50+mBERT with saved predictions, while the remaining architectures are reported from the initial runs.

\paragraph{Closed-pair constraint.} All models observe only the image--caption pair, with no access to external evidence, retrieval, knowledge bases, or metadata.

\section*{Ethical Considerations}

All data were collected from publicly available sources, and no private communications or non-public information are included. The dataset is released under a research-only non-commercial licence (CC BY-NC 4.0), which does not transfer rights in the third-party images; images are referenced by URL only and may become unavailable. Requests to remove an entry or an image link can be sent to the corresponding author; the link is disabled during review and removed if the request is valid.

\section{Conclusion}

We introduce \nepooc, the first publicly available Nepali-dominant multilingual OOC benchmark, with 1,090 image--caption pairs annotated across five typologies (Cohen's $\kappa = 0.84$) and partitioned into train/validation/test splits of 754/108/228. Systematic evaluation yields three principal observations. First, text-only mBERT and the best multimodal system (ResNet-50+mBERT) reach 94.65 and 94.56 Macro-F1 with no statistically detectable difference, which does not establish equivalence. Second, image-only models perform at chance, as expected because both captions of an item share one image, and a trained fusion model changes at most 1 of 228 predictions when its images are replaced. Third, terminal punctuation and caption length separate the classes to a large degree, and retraining on captions without punctuation lowers Macro-F1 by about 21 pp for both models, so the strong text-only scores should be read as caption discrimination on the current paired benchmark, not as evidence of semantic understanding or of the irrelevance of images. Text-only and caption-style baselines should therefore be reported alongside OOC benchmarks.

Future work should attach OOC captions to different images, balance caption style across labels, prioritise dataset expansion to 5,000--10,000 samples with broader event and source coverage, retrieval-augmented architectures to address knowledge-dependent typologies, and cross-lingual extension to neighbouring Devanagari languages to assess generalisation.

\section*{Data Availability}

\nepooc{} is publicly available under a research-only non-commercial licence, including image URLs, captions, binary and typology labels, and the predefined train/validation/test splits. Dataset: available on Hugging Face at \url{https://huggingface.co/datasets/theonlysanjeev/nepal-ooc-misinformation}. \textbf{Code and Implementation:} the benchmark evaluation code, model implementations, and preprocessing scripts are available at \url{https://github.com/SanjeevKCodes/nepooc}. The repository also contains the analysis pipeline, per-sample predictions and per-seed results of the re-run models.

\clearpage
\bibliography{references}

\appendix
\clearpage
\onecolumn

\section{Training Configuration}
\label{app:config}
\begin{table}[H]
\centering\small
\setlength{\tabcolsep}{3pt}
\begin{tabular}{@{}lllcccl@{}}
\toprule
Model & Optimiser & LR & Batch & Epochs & Pat. & Schedule \\
\midrule
CNN+LSTM & Adam & $1\!\times\!10^{-4}$ & 32 & 80 & 10 & StepLR \\
ViT+TCN & AdamW & $5\!\times\!10^{-5}$ & 32 & 100 & 10 & Cosine + warm-up \\
ResNet-50+mBERT & AdamW & $10^{-4}$ / $2\!\times\!10^{-5}$ & 32 & 50 & 10 & Cosine + warm-up \\
CLIP & AdamW & adaptive & 8 & 100 & 12 & none \\
ViT+MuRIL & AdamW & adaptive & 8 & 100 & 12 & Cosine + warm-up \\
\bottomrule
\end{tabular}
\caption{Training configuration. Weight decay: $10^{-5}$ (CNN+LSTM), $10^{-4}$ (ViT+TCN, ResNet-50+mBERT), 0.05 (CLIP, ViT+MuRIL). Adaptive learning rate: $10^{-4}$ for training fractions up to 50\%, $5\times10^{-5}$ above. StepLR: step 10, $\gamma = 0.5$. ResNet-50+mBERT learning rates: visual / text encoder. Pat.\ = early-stopping patience.}
\end{table}

\section{Caption-Only Baselines}
\label{app:base}
\begin{table}[H]
\centering\small
\begin{tabular}{@{}lcc@{}}
\toprule
 & \multicolumn{2}{c}{Macro-F1} \\
\cmidrule(l){2-3}
Caption-only model & Original & No punctuation \\
\midrule
Length LR & 68.3 & 67.3 \\
Style/punctuation LR & 92.5 & 66.0 \\
Length + style LR & 92.5 & 65.6 \\
Word TF-IDF LR & 66.8 & 66.8 \\
Word TF-IDF SVM & 64.6 & 66.4 \\
Char TF-IDF LR & 84.1 & 67.3 \\
Char TF-IDF SVM & 91.7 & 66.9 \\
\bottomrule
\end{tabular}
\caption{Caption-only baselines on the released test split ($n = 228$), with original and punctuation-free captions.}
\end{table}

\section{Publisher Hold-out}
\label{app:pub}
\begin{table}[H]
\centering\small
\setlength{\tabcolsep}{4pt}
\begin{tabular}{@{}llcccc@{}}
\toprule
Held-out publisher (rows) & Captions & Char SVM & Style LR & Length LR & Word LR \\
\midrule
\multirow{2}{*}{\texttt{nepalfactcheck.org} (508)} & Original & 93.5 & 98.8 & 75.9 & 59.9 \\
 & No punct. & 62.0 & 73.8 & 82.2 & 60.0 \\
\midrule
\multirow{2}{*}{\texttt{techpana.com} (382)} & Original & 78.8 & 96.3 & 38.0 & 57.1 \\
 & No punct. & 61.9 & 40.1 & 37.9 & 58.0 \\
\bottomrule
\end{tabular}
\caption{Macro-F1 when training on all other publishers and testing on the held-out publisher. This is a diagnostic of source dependence, not external validation.}
\end{table}

\section{Language-wise Results}
\label{app:lang}
\begin{table}[H]
\centering\small
\begin{tabular}{@{}lccc@{}}
\toprule
Caption-only model & Nepali (180) & English (34) & Code-sw. (14) \\
\midrule
Length LR & 65.6 & 65.2 & n/a \\
Style/punctuation LR & 96.1 & 71.5 & n/a \\
Word TF-IDF LR & 63.7 & 79.0 & n/a \\
Char TF-IDF SVM & 94.4 & 76.1 & n/a \\
\bottomrule
\end{tabular}
\caption{Test Macro-F1 by caption language (number of test rows in parentheses). The English and code-switched groups contain 17 and 7 pairs and are too small for firm conclusions; n/a marks groups where single-class predictions leave Macro-F1 undefined.}
\end{table}

\section{Typology Details}
\label{app:typ}
\begin{table}[H]
\centering\small
\begin{tabular}{@{}lcccc@{}}
\toprule
OOC typology (five-way) & Support & P & R & F1 \\
\midrule
Fabricated & 62 & 0.75 & 0.87 & 0.80 \\
Miscaptioned & 28 & 0.55 & 0.57 & 0.54 \\
Temporal mismatch & 13 & 0.97 & 0.46 & 0.62 \\
Geographic mismatch & 9 & 0.81 & 0.49 & 0.59 \\
Identity mismatch & 2 & 0.00 & 0.00 & 0.00 \\
\midrule
Macro-F1 (5 seeds) & 114 & \multicolumn{3}{c}{$50.95 \pm 3.80$} \\
\bottomrule
\end{tabular}
\caption{Per-class results of fine-tuned mBERT on the OOC test captions (mean over 5 seeds). Identity mismatch has two test captions, so results are exploratory.}
\end{table}

\section{OOC-Class Precision and Recall}
\label{app:ooc}
\begin{table}[H]
\centering\small
\setlength{\tabcolsep}{3pt}
\begin{tabular}{@{}lcccc@{}}
\toprule
Model & Macro-F1 & OOC P & OOC R & OOC F1 \\
\midrule
CNN+LSTM & 78.15 $\pm$ 10.97 & 79.23 $\pm$ 11.27 & 76.32 $\pm$ 11.42 & 77.72 $\pm$ 11.19 \\
ViT+TCN & 92.10 $\pm$ 1.36 & 92.20 $\pm$ 2.81 & 92.11 $\pm$ 1.96 & 92.11 $\pm$ 1.24 \\
CLIP & 69.00 $\pm$ 0.72 & 74.97 $\pm$ 1.34 & 58.25 $\pm$ 1.33 & 65.54 $\pm$ 0.87 \\
ResNet-50+mBERT & 94.56 $\pm$ 0.24 & 95.36 $\pm$ 0.37 & 93.68 $\pm$ 0.39 & 94.51 $\pm$ 0.24 \\
ViT+MuRIL & 93.33 $\pm$ 0.37 & 94.14 $\pm$ 1.22 & 92.46 $\pm$ 1.71 & 93.27 $\pm$ 0.43 \\
\bottomrule
\end{tabular}
\caption{OOC-class precision (P), recall (R) and F1 at 100\% training data (mean $\pm$ std over 5 seeds).}
\label{tab:ooc}
\end{table}

\section{Exact McNemar Test per Seed}
\label{app:mcn}
\begin{table}[H]
\centering\small
\begin{tabular}{@{}lcccc@{}}
\toprule
 & \multicolumn{2}{c}{Original captions} & \multicolumn{2}{c}{No punctuation} \\
\cmidrule(lr){2-3}\cmidrule(l){4-5}
Seed & Discordant & Exact $p$ & Discordant & Exact $p$ \\
\midrule
42 & 0 & 1.000 & 20 & 0.503 \\
123 & 1 & 1.000 & 25 & 0.230 \\
456 & 1 & 1.000 & 26 & 1.000 \\
789 & 1 & 1.000 & 21 & 0.383 \\
2024 & 0 & 1.000 & 19 & 1.000 \\
\bottomrule
\end{tabular}
\caption{Exact (binomial) McNemar test between mBERT and ResNet-50+mBERT on matched test samples (228 rows per seed).}
\end{table}

\section{Integrity Audit}
\label{app:integrity}
\begin{table}[H]
\centering\small
\begin{tabular}{@{}lr@{}}
\toprule
Check on the released CSV files & Result \\
\midrule
\texttt{post\_id}s / rows & 545 / 1,090 \\
\texttt{post\_id}s without one Pristine and one OOC row & 0 \\
Pairs with different image URLs & 0 \\
\texttt{post\_id}s in more than one split & 0 \\
Image URLs in more than one split & 0 \\
Unique image URLs (545 \texttt{post\_id}s) & 527 \\
Pairs with identical captions (exact) & 6 \\
Pairs with identical captions after normalisation$^{\dagger}$ & 27 \\
\quad of which in the test set (114 pairs) & 4 \\
Rows labelled Nepali whose captions are mostly Latin script & 38 \\
Groups of identical image files (SHA-256) / across splits & 9 / 0 \\
Perceptual-hash candidate pairs / across splits & 27 / 13 \\
\quad confirmed same photograph across splits & 1 \\
\bottomrule
\end{tabular}
\caption{Integrity audit. $^{\dagger}$Punctuation and whitespace removed, lower-cased.}
\end{table}

\end{document}